\documentclass[12pt]{article}
\usepackage{times}
\usepackage{comment}
\usepackage{xcolor}
\usepackage{amsmath}
\usepackage{amsmath,amssymb,amsfonts}
\usepackage{amsbsy}
\usepackage{graphicx}
\usepackage{siunitx}

\newenvironment{sciabstract}{%
\begin{quote} \bf}
{\end{quote}}

\title{Overcoming the Force Limitations of Magnetic Robotic Surgery: Impact-based Tetherless Suturing} 

\author
{Onder Erin,$^{1}$ Xiaolong Liu,$^{1}$ Jiawei Ge,$^{1}$ Lamar Mair,$^{2}$\\ Yotam Barnoy,$^{3}$ Yancy Diaz-Mercado,$^{4}$ Axel Krieger,$^{1*}$\\
\\
\normalsize{$^{1}$Department of Mechanical Engineering, Johns Hopkins University, Baltimore, MD 21211 USA}\\
\normalsize{$^{2}$ Weinberg Medical Physics, Inc., North Bethesda, MD 20852 USA}\\
\normalsize{$^{3}$ Department of  Computer  Science,  Johns  Hopkins  University,  Baltimore,  MD  21211 USA}\\
\normalsize{$^{4}$ Department  of  Mechanical  Engineering,  University  of  Maryland,  College Park, MD 20742 USA}
\\
\normalsize{$^\ast$E-mail: axel@jhu.edu.}
}

\date{}

\begin{document}

\maketitle 
\begin{sciabstract}
  Magnetic robotics obviate the physical connections between the actuators and end effectors resulting in ultra-minimally invasive surgeries. Even though such a wireless actuation method is highly advantageous in medical applications, the trade-off between the applied force and miniature magnetic end effector dimensions has been one of the main challenges in practical applications in clinically relevant conditions. This trade-off is crucial for applications where in-tissue penetration is required (e.g., needle access, biopsy, and suturing). To increase the forces of such magnetic miniature end effectors to practically useful levels, we propose an impact-force-based suturing needle that is capable of penetrating into \textit{in-vitro} and \textit{ex-vivo} samples with 3-DoF planar freedom (planar positioning and in-plane orienting). The proposed optimized design is a custom-built 12 G needle that can generate 1.16 N penetration force which is 56 times stronger than its magnetic counterparts with the same size without such an impact force. By containing the fast moving permanent magnet within the needle in a confined tubular structure, the movement of the overall needle remains slow and easily controllable. The achieved force is in the range of tissue penetration limits  allowing the needle to be able to penetrate through tissues to follow a suturing method in a teleoperated fashion. We demonstrated \textit{in-vitro} needle penetration into a bacon strip and successful suturing of a gauze mesh onto an agar gel mimicking a hernia repair procedure.\\
  \\
  Summary: First implementation of magnetic suturing with an impact-based magnetic needle.
\end{sciabstract}
\baselineskip24pt

\paragraph*{Main Text}\mbox{}
\paragraph*{INTRODUCTION}\mbox{} \\
Magnetically guided robotic systems could revolutionize surgery. State-of-the-art surgical robots (e.g., da Vinci Surgical System \cite{maeso2010efficacy}) enable minimally invasive procedures by providing surgeons with finer control and increased mobility via a few small incisions.
However, these robotic manipulators have large mechanical footprints in the patient compared to the surgical end-effectors (e.g., needles and grippers), making the robotic manipulator the most invasive component. 
Alternatively, magnetic fields could wirelessly guide magnetic agents (i.e., end-effectors) accurately and safely, obviating the mechanical manipulator and paving the way for untethered robotic surgical systems  \cite{abbott2020magnetic, erin2019, kummer2010octomag,  martel2007automatic, erin2017design}. Such a system would enable ultra-minimally invasive procedures, like the closing of a hole in the heart or repairing a hernia with just a pin prick for access.

Size of the tools is a crucial factor that may determine the minimally invasiveness of surgeries \cite{nelson2010microrobots, chen2018small, sitti2015biomedical, simaan2018medical, huda2016robots}. Even though these magnetic end effectors are miniaturized with no need to embody an on-board battery source, electronics and circuitry \cite{mahoney2016five, ciuti2010robotic, carpi2010magnetically, yim2011design}, the magnetic pulling forces scales with $L^3$ ($L$ is the characteristic length of the magnetic object), making the magnetic pulling forces not favorable at small scales. Clinically relevant applications that include tissue penetration, biopsy, or suturing become either technically demanding or infeasible in such scales. Required forces can either be acquired by bringing the electromagnetic coils to a very close vicinity of the magnet \cite{son2019magnetically, kobayashi2008novel}, or increasing the power of the magnets with larger and larger coil currents, which in turn creates difficulties due to resistive losses and heating \cite{leclerc2018magnetic, rahmer2018remote}. In many cases, both approaches are applied to compromise between the sweet spot of minimally invasive dimensions and actuation system capabilities. 

Magnetically actuated biopsy procedures have been demonstrated in \textit{in-vitro} and \textit{ex-vivo} conditions. Due to the high force requirements of these systems, unorthodox robot design methodologies have been developed. Vartholomeos et. al has demonstrated a grounded gear train that is powered by the magnetic pulling forces generated inside of an MRI device \cite{vartholomeos2013mri}. The grounded robotic system is on the order of 10s of cm in size located outside of the patient. The biopsy needle can be pushed and pulled along a single desired direction in 3D space. In order to meet the high force requirements, this system uses a lever arm and a rotationary mechanism connected to a gear train mechanism that results in a translational motion for the needle tip. Even though this system can be useful to acquire biopsy samples from surface-level regions, this system can not be used for minimally invasive suturing operations deep inside the tissue. As an alternative approach, untethered magnetic capsule-based biopsy operations can be performed only at the near vicinity of the electromagnetic actuators due to the high force requirements \cite{son2019magnetically}. Such systems would be undesired for suturing applications with a thread because the suturing location may not necessarily be near the electromagnets. More importantly, the diameter of the capsule is much larger compared to the diameter of the needle, which makes a full-body penetration infeasible.

In recent studies, researches have investigated the advantage of impact forces to generate large instantaneous forces, which meets the high force requirements in various applications \cite{leclerc2017magnetic, quelin2021experimental}. The fundamental principle in these studies is to allow the magnetic element to gain a large enough velocity, which will create a strong momentarily force under a collision event as a result of a momentum transfer in a short period of time. This technique has been used by Quelin et. al to create a two-dimensional locomotion of a microrobot prototype that is impact-driven. Similarly, Leclerc et. al has demonstrated usage of such an impact force for developing a single-directional magnetic hammer. A 6-mm in diameter sphere magnet is being pulled and pushed inside a hollow tube to penetrate into an \textit{ex-vivo} lamb brain. However, considering the large size of the capsule, and incapability of orienting into any desired penetration angle, such a capsule hammer could be used for biopsy along a single axis towards a single direction but could not be used for minimally invasive suturing applications. 

Even though a biopsy task has a requirement for a partial tissue penetration, the level of maneuverability of a suturing task could not be met in these aforementioned studies. Therefore, although utilizing such impact forces are promising, there is no viable magnetic robot that can accomplish suturing in a tissue. The magnetosuture system studied by Mair et. al \cite{mair2020magnetosuture} operates inside of liquid environments and the static average forces generated by the system is far from the clinical force requirements for a high degrees-of-motion suturing capability. 

\begin{figure}
    \centering
    \includegraphics[width=\textwidth]{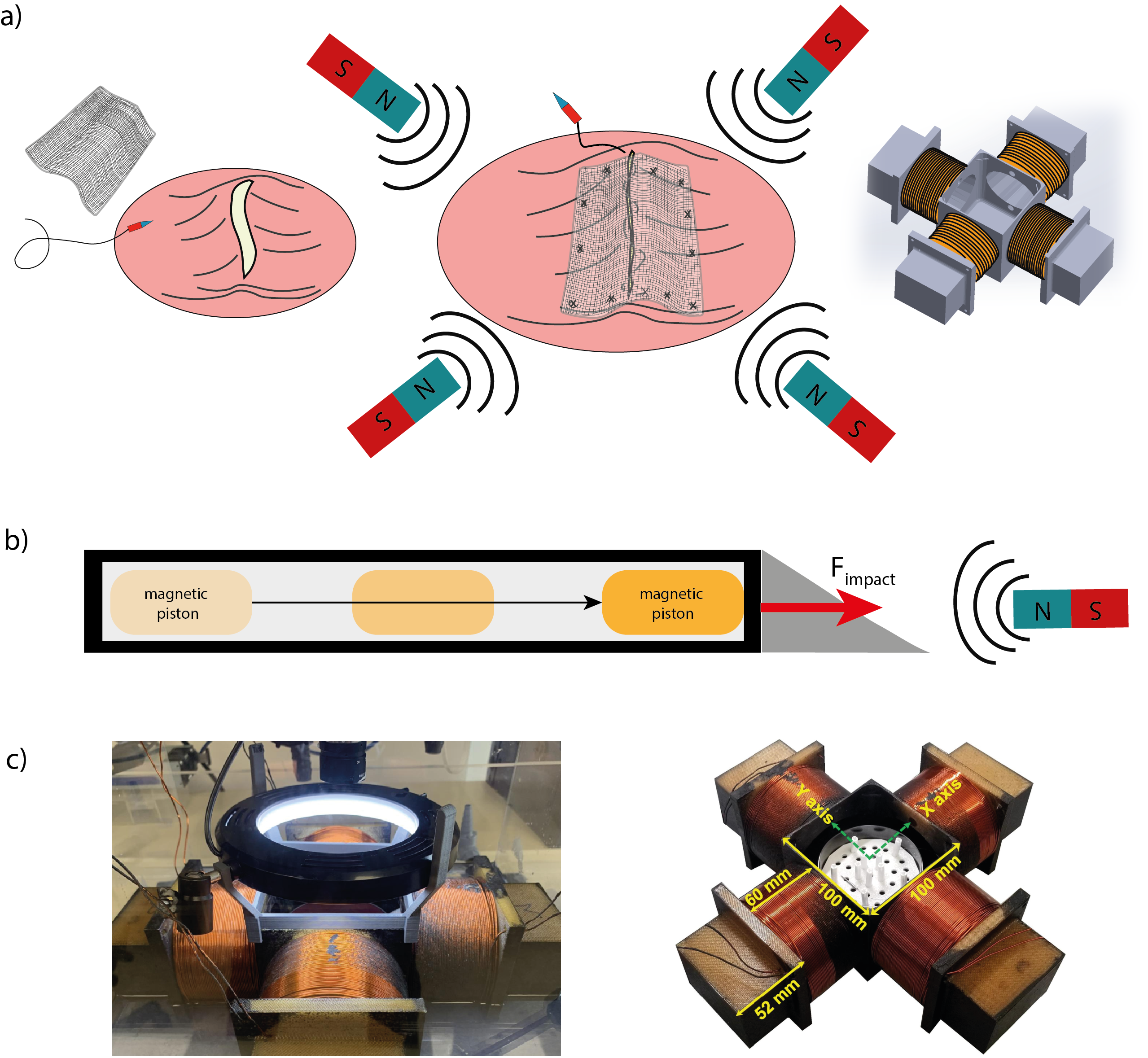}
    \caption{Magnetic suturing. a) A typical procedure for hernia repair is to use a mesh and a suturing needle to close a defect. A magnetic needle controlled by external magnetic fields could accomplish this procedure remotely by revolutionizing the surgery with this ultra minimally invasive approach. b) To overcome the force limitation for miniature magnetic robots, we designed and manufactured an impact-based needle, which utilizes a moving magnetic piston's momentum to realize momentarily high force outputs for tissue penetration. c) The running suture path to stitch a mesh into an agar gel is accomplished by the four electromagnetic coil system. }
    \label{fig:concept}
\end{figure}

Here in this study, we present an impact-based magnetic needle that can accomplish 56 times stronger penetration forces compared to the standard magnetic pulling techniques without any size or electromagnetic power increase. We utilize the electromagnetic system and the needle design towards clinical magnetic suturing tasks as depicted in Fig. \ref{fig:concept}. We designed and custom-built a 12 G needle with a tubular body that contains a piston magnet. Since the magnetic piston is contained within a constrained volume, the penetration operation also remains safer compared to such a collision events in a free-space. Additionally, we present the design optimization that maximizes the impact force generated at the needle tip by guiding the selection of the magnetic piston length and the needle length. With this optimized design, we demonstrated penetration into \textit{ex-vivo} tissues. Moreover, we demonstrated a running suture implementation with \textit{in-vitro} agarose gel samples towards clinically relevant suturing tasks.

\begin{figure}
\centering
	\includegraphics[width=0.8\textwidth]{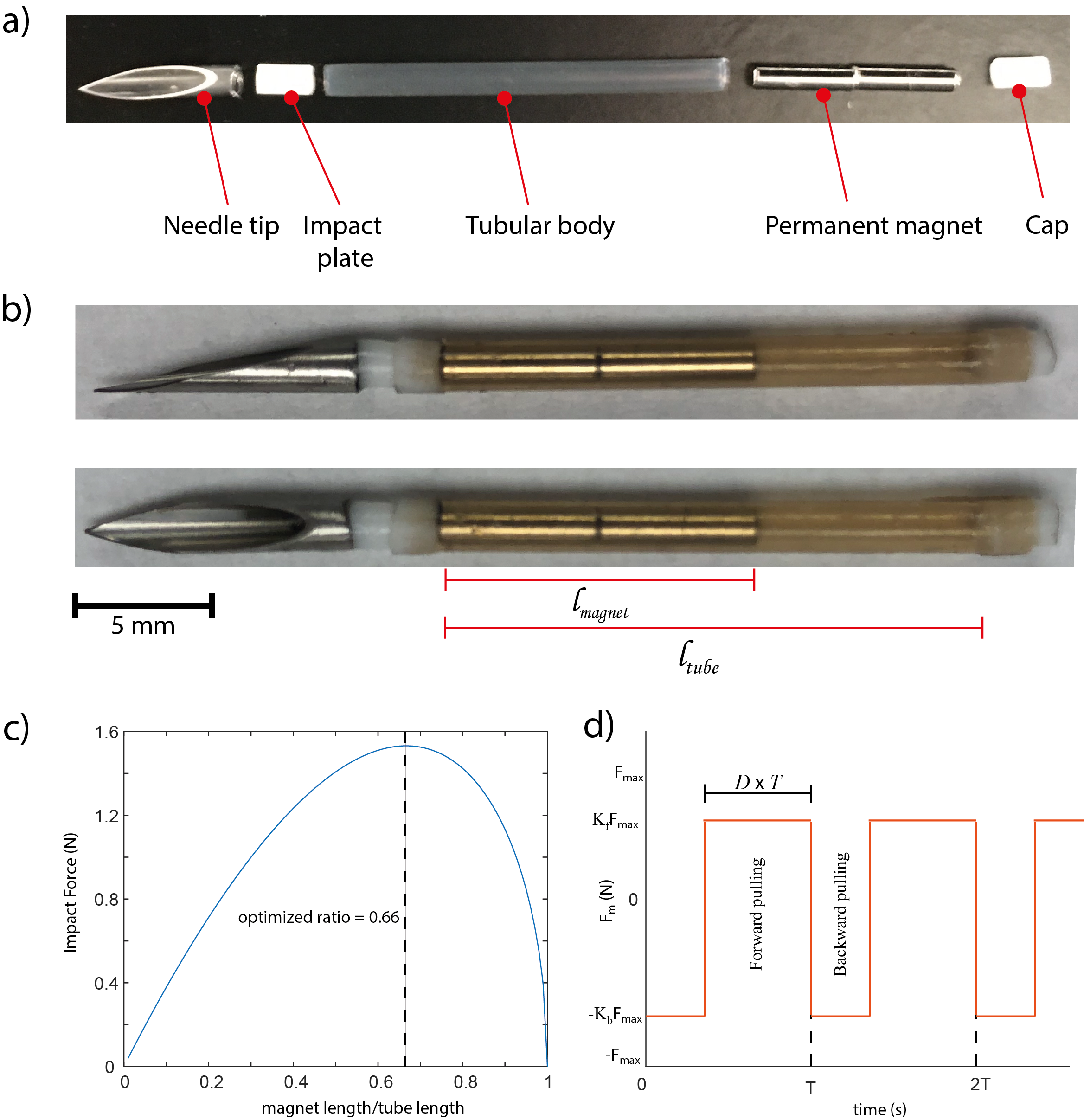}
	\caption{Impact-based needle design, magnet length optimization, and actuation sequence optimization. a) The impact-based magnetic needle consists of five main components: the needle tip, impact plate, tubular body, permanent magnet, and a cap. b) These components are assembled via cyanoacrylate-based adhesives or press-fit inside the tubular structure. The permanent magnet is slightly smaller than the diameter of the tubular structure and it can freely move back and forth within the tube. c) Increasing the length of the magnet increases the applied magnetic force but reduces the possible travelling distance within a limited tubular body. To maximize the impact force, the optimum size of the magnet is found to be 0.66 times of the overall tubular body size. d) The motile magnet is being pulled back and forth in the tubular body. A period is demonstrated by $T$, forward pulling cycle ratio is demonstrated by $D$. $K_f$ and $K_b$ are the coefficients for the magnitude of the pulling forces. Depending on the parameters of $D$, $T$, $K_f$ and $K_b$, the penetration forces vary. }
	\label{fig:design}
\end{figure}

\paragraph*{RESULTS}
\paragraph*{Robot Design}\mbox{} \\
The proposed robot in this study embodies the features of a suturing needle while incorporating a dynamically moving magnetic piston inside the needle. The overall robot shown in Fig.~\ref{fig:design}a) is comprised of five components including 1) a needle tip, 2) a tubular body, 3) a permanent magnet, 4) an impact plate, and 5) a cap. In order to keep the dimensions small while demonstrating the suturing concept, we selected a 12 G needle size as a nice compromise between the manual manufacturability of the needle and relevance to the medical applications. The permanent magnet is a NdFeB magnet with 1.59 mm in diameter and 12.7 mm in length. The 19.05 mm long tubular body is made out of PTFE plastic with an outer diameter of 2.08 mm and inner diameter of 1.67 mm. The impact plate and cap is composed of a glass-mica cylindrical rod. The magnet diameter is selected as slightly smaller than the inner tube diameter to use the space efficiently for delivering enough force. 

The length of the magnet and the tubular body are selected based on an optimized length. The impact force dynamics and the resultant optimization process is described in the Materials and Methods Section. The impact plate is responsible to deliver the impact force to the needle tip as efficiently as possible. Therefore, a glass-mica ceramic is one of the stiffest non-conductive machinable alternatives. The needle tip is cut out from a regular 12 G stainless steel needle. In order to prevent deformations during the cutting with pliers, a sacrificial metal rod is inserted inside the needle to provide support. Both the impact plate and cap are assembled with a press fit to the plastic tubular body. The needle tip is attached to the impact plate by using cyanoacrylate-based adhesives. A similar adhesive is used for attaching the suture thread at the back of the needle.

\paragraph*{Impact Force Characterization}\mbox{} \\
The magnetic actuation sequence of backward and forward motion creates the impact force needed for tissue penetration. The performance of the overall impact force depends on the period of the sequence, $T$, duty ratio of the forward motion, $D$, forward and backward pulling force constants, $K_f$ and $K_b$. For an ideal impact force generation, it is desired to have the strongest pulling force applied (i.e., $K_f = 1$), the magnet located at the tail travels all the distance along the tube and hits to the impact plate until the momentum of the magnet is completely transferred. This means the selection of both $D$ and $T$ determines the performance of the impact penetration. During the cyclic sequence, if these parameters are not chosen properly, the magnet does not transfer the momentum completely, or the magnet would not be able to reach to the impact plate to transfer the momentum at all. Similarly, backward force also has to be tuned properly such that the magnet travels backward and regains the travelling distance for the next impact motion. 

\begin{figure}
\centering
	\includegraphics[width=0.8\textwidth]{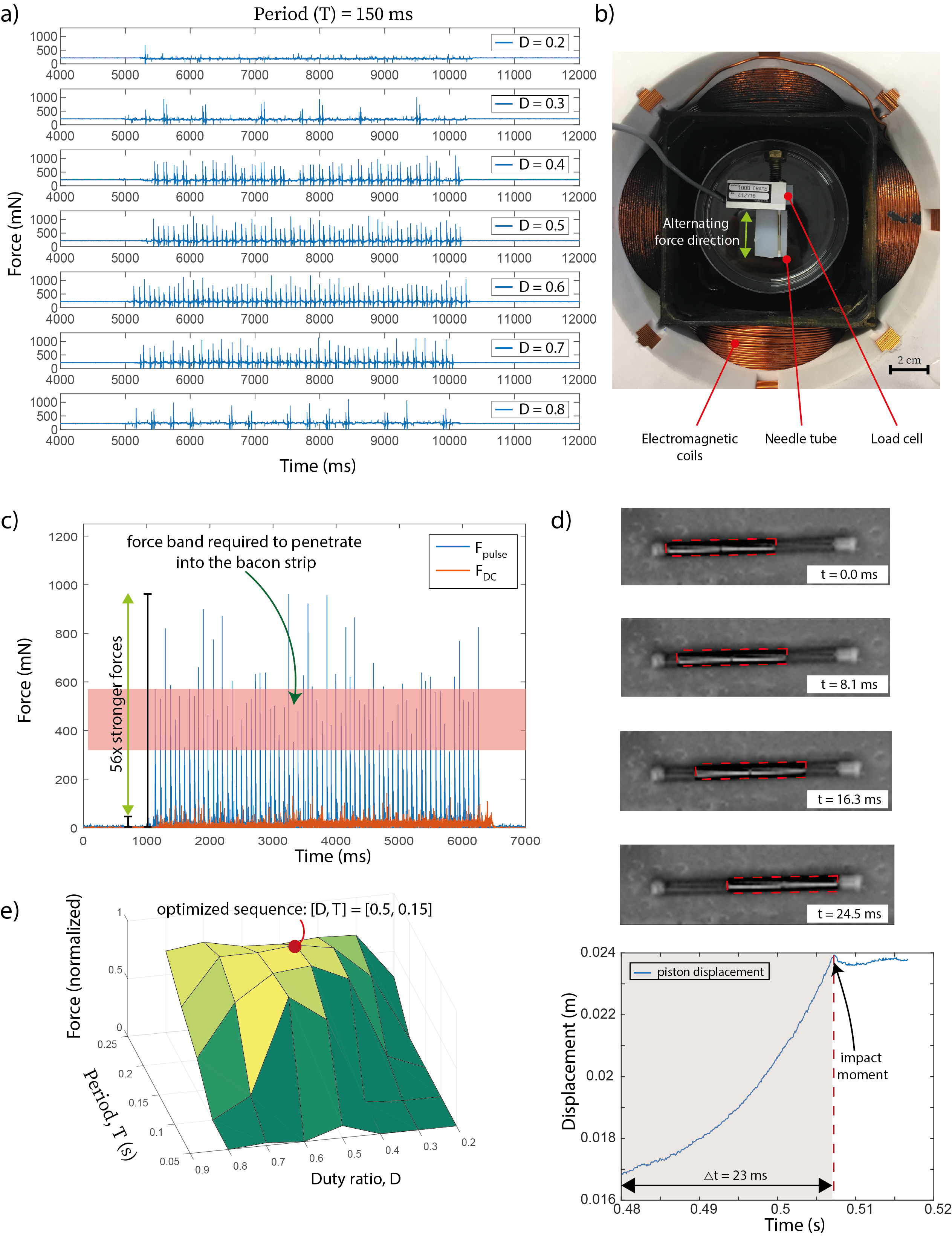}
	\caption{Characterization experiments. a) For a selected well-performing period duration, $T$ = 150 ms, the impact force measurements with respect to time are shown. The duty ratio, $D$, determines the forward pulling duration and therefore the impact motion is affected by the parameter. Values ranging from 0.2 to 0.8 show the force output of the needle. Having D in the range of 0.4 to 0.6 yields more than 1000 mN momentarily forces in the needle. b) The forces are measured using a load cell under the region of interest of the electromagnetic coil system. c) Compared to the DC pulling force, the impact-based mechanism provides 56 times higher forces to allow penetration into tissue. d) The velocity of the magnetic piston is being tracked with a high speed camera at 15000 fps. e) Considering the various span of $D$ and $T$ variables, the force ratings are shown in the figure. Selection of impractical $D$ and $T$ values  results in degradation of the force performance. The optimal value for $D$ and $T$ is found to be 0.5 and 0.15 s for this study. }
	\label{fig:characterization}
\end{figure}

In our experimental results, as presented in Fig. \ref{fig:characterization}, the ideal period of the sequence, $T$, was found to be 0.15 s while $D$ is 0.5 when the needle is at the center of the workspace. Therefore, these values of $T$ and $D$ are being used for for the force characterization measurements.

The data for force characterization is acquired at two different locations: 1) at the center, 2) at the far end. At each locations, both pulsing forces to see the impact force and DC pulling force are measured. Each measurements are repeated 3 times. The acquired data is used to compare the impact based penetration force with the typical magnetic pulling force without any impact for the same needle. This comparison has shown that for the same needle, while the impact-force based mechanism can provide up to 1163 mN of impact force, the continuous pulling force can only provide 18 mN of force at the center. This is the force estimate for bacon strip penetration experiments, where the magnetic piston remains around the center. Similarly, for the measurements at the far end, average of 610 mN impact force is provided while the DC pulling force can provide only 12 mN of force.This measurement at the far end is relevant to the experiments with agar gel with gauze penetrations since the magnetic piston remains around the far end. Such a large increase in the force with a design modification allows us to penetrate tissues by using standard magnetic field systems.
These force measurement results are provided in Fig. \ref{fig:characterization}.

\paragraph*{Suturing with Magnetic Impact Needles}\mbox{} \\
The purpose of this study is to utilize the benefit of a high force for a challenging tissue penetration. Due to the scaling law for magnetic actuation, it is challenging to have high enough forces for penetration with a miniature needle. For a tetherless suturing task, it is a must to demonstrate proper suturing. In order to demonstrate a basic suturing with a mesh, and to penetrate \textit{in-vitro} samples, we implemented two demonstration experiments. 

In the first experiment, we clamped a single bacon strip with 1.6 mm in thickness. The cross-section of the bacon strip is held perpendicular to the planar workspace. While monitoring the needle and the sample with an optical camera, we performed needle steering, locomotion and a successful penetration of the needle into the bacon strip. The back-and-forth hammering motion is automated while the user provides the direction of the penetration and tunes the intensity of the backward pulling force. To match the experimentally measured force values with this experiment, the magnetic piston has to be at the center. Considering the length of the needle, therefore, we had to put the bacon strip towards the edge of the workspace. Due to the space constraints, such a configuration allowed a successful penetration but we could not steer the magnet for a full suturing routine due to the insufficient workspace. However, this experiment has shown the great promise of an impact-force mechanism for the future of the tetherless suturing tasks. Overall penetration duration has been 16 s and a total of 11.8 mm of the needle has penetrated through the tissue. 

\begin{figure}
	\includegraphics[width=1\textwidth]{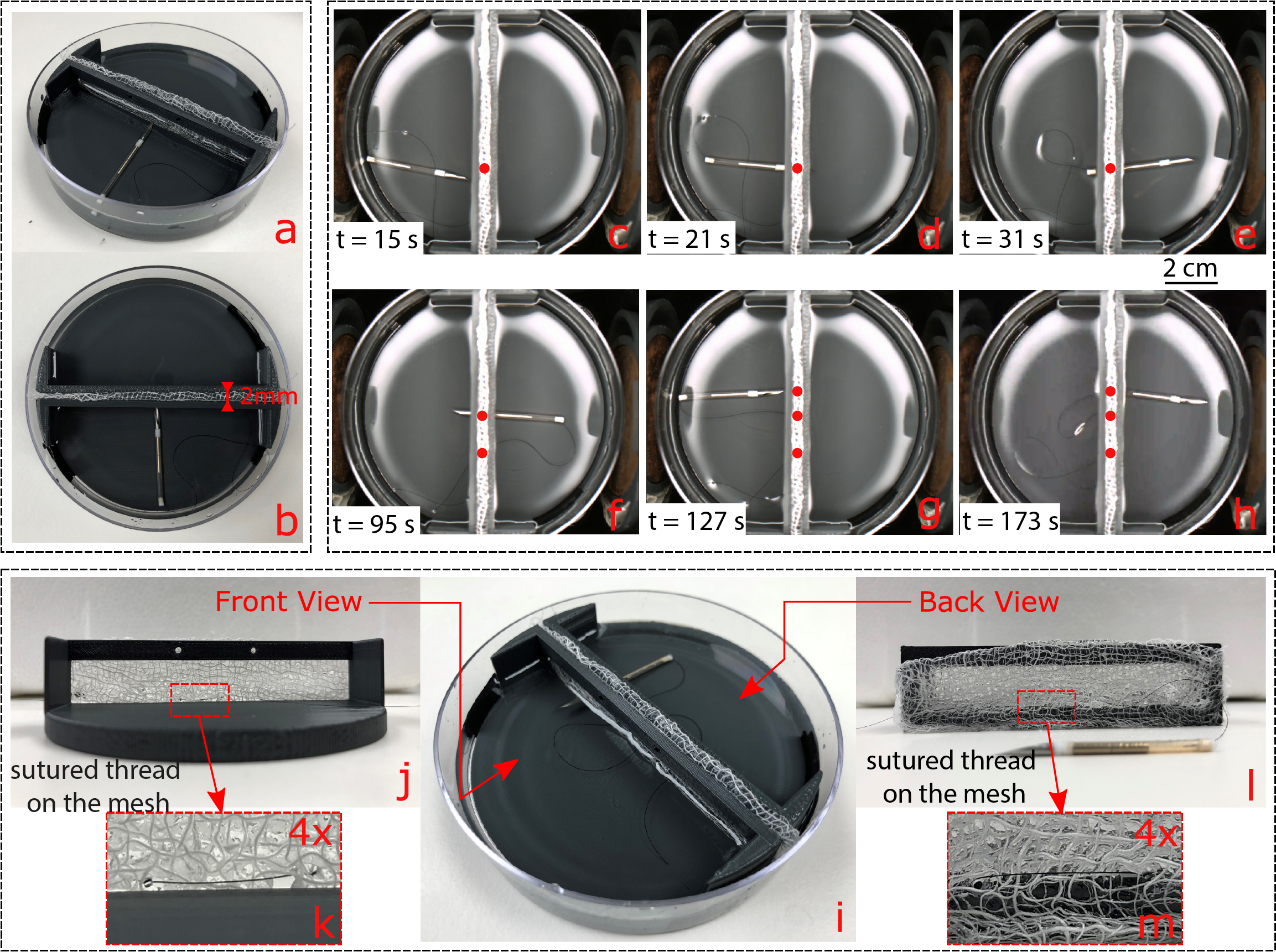}
	\caption{Suturing on agarose gel. a) The experimental setup for demonstrating the suturing capacity on an agar gel: we prepared a clamp mechanism that holds the sample and allows mobility for the suturing needle. b) The agar gel is 0.6\% and 2 mm in thickness. A gauze mesh is covered around the agar gel to represent the meshes being used in hernia repair. c-h) The needle is being steered by a joystick in an open-loop fashion. An overhead camera is being used to provide real-time monitoring of the workspace. The needle has demonstrated three penetrations in less than 3 minutes. j-m) The suture thread used for suturing the mesh and the agar gel is shown after the completion of the suturing task. The suture thread used for the experiments is 50 $\mu$m thick.}
	\label{fig:gel}
\end{figure}

In the second experiment, we implemented a suturing task with a gauze mesh to mimic a hernia repair. 
We prepared 0.6\% agarose gel with 3 mm thickness and covered with a gauze mesh \cite{pervin2011mechanically}. This sample is clamped vertically and located along the centerline of the workspace. As mentioned above, such a configuration allows us to steer the needle freely to demonstrate the suturing tasks within the workspace of our magnetic system.
The suturing was performed in a teleoperated manner with commands from a joystick mapping to coil currents. The applied running suture, consisting of 3 penetrations (2 from the front, 1 from the back) has taken 158 s to complete. It is important to note that occasional torquing the needle by applying a magnetic force perpendicular to the direction of the needle penetration helps for suturing, especially when the magnetic piston is quite far from the electromagnets. Results of these experiments with experimental images are demonstrated in Fig. \ref{fig:gel}.

\paragraph*{DISCUSSION}\mbox{} \\
One of the key challenges in magnetically actuated robotics is the scaling law of the applied magnetic forces at miniature scales. Miniature robots either require special designs for a true penetration into a tissue, or they require very high electrical power and short distances to electromagnets that are incompatible with clinically relevant applications. In this work, we are demonstrating impact-force based suturing that can boost the penetration forces by a factor of 56, allowing the needles to penetrate into tissue-like structures successfully. This mechanism paves the way for suturing tasks in clinically relevant settings, where repetitive mechanical penetration is required. 

Combining such a needle with a more powerful electromagnetic coil system could provide even stronger forces to accomplish the tasks faster, allow further needle miniaturization, and provide sufficient forces at longer distances. Such a strong setup requires active water cooling and more power. The electromagnetic coils might also increase in size which increases the response time of the electromagnetic coils. The frequency range of operation (up to 15 Hz) could still be provided with no issue even for larger electromagnets. 

The level of autonomy proposed in this study is commanding the suturing needle via the user's joystick inputs. Because of the nonlinearity of the magnetic field and the rapid response of the needle, precise controlling with teleoperation is challenging. One of the tasks that is open for improvement is working on assistive or fully autonomous control of the needle under the presence of impact dynamics. Such assistive needle control could take the user input as the desired position and orientation and autonomously steer the needle into this desired position. A fully autonomous suturing needle would need to be capable of assessing and analyzing the overall suturing state. This requires advanced image processing algorithms as well as effective decision making and needle steering strategies \cite{yang2018grand}.

The needle size used in this study is 12 G, which is at the large end of possible suturing needles. Even though such a size is capable of demonstrating the proof-of-concept and the advantage of the impact mechanism, further miniaturization of the needle will be required for \textit{in-vivo} applications. With precise manufacturing techniques, the needle size can be made smaller. Combining such a needle with an improved electromagnetic coil system would allow \textit{in-vivo} applications with large workspaces. 

Lastly, proper imaging methodologies for \textit{in-vivo} experiments should be integrated for autonomous tasks. Such imaging methodologies can consist of laparoscopic optical cameras, ultrasound, MRI, or X-ray fluoroscopy. The study presented here is the closest to using laparoscopic optical images. We envision that \textit{in-vivo} experiments would add additional complexity with respect to coordinate fixing, 3-D localization, and localization challenges in dynamic and cluttered surgical environments. These challenges are also applicable for any tetherless magnetic end-effectors,and solving these challenges would bring magnetic biomedical applications one step closer to the clinical settings.

\paragraph*{MATERIALS AND METHODS}
\paragraph*{Magnetic Actuation and Impact Dynamics}\mbox{} \\
The magnetic pulling forces, $\mathbf{F}_m$, and torques, $\mathbf{T}_m$, acting on a magnetic body can be presented as in the equations below:
\begin{gather}
    \mathbf{T}_m = V_m(\mathbf{M}_m\times \mathbf{B}) \label{eqn:magnetic_torque},\\
    \mathbf{F}_m = V_m(\mathbf{M}_m\cdot \nabla)\mathbf{B}\label{eqn:magnetic_force},
\end{gather}

\noindent where $V_m$ is the volume, $\mathbf{M}_m$ is the average magnetization of the magnetic piston, and $\mathbf{B}$ is the magnetic field vector generated by the electromagnetic coils. $\mathbf{T}_m$ allows rotation of the needle and pointing the needle tip into any desired location. $\mathbf{F}_m$ provides the pulling force on the magnetic piston causing the translation and hammering motion inside the tube.

For the magnetic piston travelling from the tail to the tip of the needle under a constant average pulling force, $\mathbf{F}_{m}$, the Euler’s laws of rigid-body motion for the magnetic piston yields the equations of dynamics as follows:
\begin{equation}
    \mathbf{F}_{net} = m\mathbf{a} = \mathbf{F}_{m} + \mathbf{F}_{impact},
    \label{eqn:force_eqn}
\end{equation}
\noindent{} where $\mathbf{F}_{net}$ is the net force on the magnetic piston and
\begin{equation}
    \mathbf{F}_{impact} = 
    \begin{cases}
    \mathbf{f}_{impact}(\mathbf{V}_{rel}) ,& \text{if } (||x_{m}||\geq x_{crit})\\
    0,              & \text{otherwise}.
    \end{cases}
    \label{eqn:force_impact}
\end{equation}

\noindent $x_{m}$ is the center location of the magnetic piston, $x_{crit}$ is the critical magnet center position where the contact to impact plate takes place, and $\mathbf{V}_{rel}$ is the relative velocity between the magnetic piston and the impact plate. $\mathbf{f}_{impact}$ can be formulated by the simplified impact physics for two rigid body collision with a constant deceleration \cite{Yankelevsky2020}
\begin{equation}
    \mathbf{f}_{impact}(\mathbf{V}_{rel}) = -\frac{m\mathbf{V}_{rel}}{\Delta{}t_{impact}},
    \label{eqn:impact_basic}
\end{equation}



As it can be seen in Eqs. (\ref{eqn:magnetic_force}), (\ref{eqn:force_eqn}), and (\ref{eqn:force_impact}), there are two types of forces acting on the magnetic piston. While $\mathbf{F}_{m}$ is the standard magnetic pulling force, $\mathbf{F}_{impact}$ occurs under the presence of contact with the impact plate.

The final force on the needle tip is the most important parameter that would maximize the penetration and suturing efficacy. Considering the movable magnetic piston inside the needle, inner tube friction and outer interactions, the torque and force acting on the needle can be represented as
\begin{gather}
    \mathbf{T}_{needle} =\mathbf{T}_m + \mathbf{T}_d \label{eqn:needle_torque},\\
    \mathbf{F}_{needle} = 
    \begin{cases} \mathbf{F}_{m} - \mathbf{F}_{impact} + \mathbf{F}_{f_i} + \mathbf{F}_d ,& \text{if } (||x_{m}||\geq x_{crit})\\
    \mathbf{F}_{m} + \mathbf{F}_{f_i} + \mathbf{F}_d,              & \text{otherwise}.
    \end{cases}\label{eqn:needle_force}
\end{gather}

\noindent External forces and torques due to the interactions with environment or tissue are represented by $\mathbf{F}_d$ and $\mathbf{T}_d$, respectively. $\mathbf{F}_{f_i}$ is the internal friction force between the motile piston and the needle tube.

\paragraph*{Impact-based Actuation Mechanism}\mbox{} \\
The impact-based actuation methodology requires an actuation sequence that pulls and pushes the magnetic piston back and forth along the direction of the penetration. The momentum accumulated on the piston travelling along the tubular structure is transferred on the impact plate in a short time causing a much stronger momentary force. This hammering behaviour provides a large penetration force on the magnetic needle that would not be achievable without a significant upgrade to the electromagnetic actuation hardware. This impact-based actuation mechanism is depicted in Fig. \ref{fig:concept}b.

The back and forth motion of the magnet creates a certain sequence for the magnetic field and the gradient field applied. For a magnet penetrating along a desired direction $\mathbf{r}$, the applied magnetic field and gradient as a function of time should have the following properties:
\begin{gather}
    \mathbf{B}_m || \mathbf{r} \label{eqn:field_requirement_1},\\
    \mathbf{F}_{m}(t) = 
    \begin{cases} K_f||\mathbf{F}_{max}||\hat{\mathbf{r}} ,& \text{if } (\bmod{(t,T)} \geq D \cdot T)\\
    -K_b||\mathbf{F}_{max}||\hat{\mathbf{r}},              & \text{if } (\bmod{(t,T)} < D \cdot T),
    \end{cases}\label{eqn:field_requirement_2}
\end{gather}
\noindent where $\times$ denotes cross product, $K_f$ and $K_b$ are the forward and backward pulling force constants, respectively, limited between 0 and 1. $\hat{\mathbf{r}}$ is the unit vector along the direction of the penetration, $\mathbf{F}_{max}$ is the maximum pulling force that is exerted on the magnet, $t$ is the current time, $T$ is the period duration for the hammering motion, $D$ is the duty ratio for the forward pulling duration limited between 0 and 1. This pulling force sequence over time with relevant parameters are depicted in Fig. \ref{fig:design}d.

In addition to the actuation parameters, the location of the needle has an important contribution. Because the magnetic field generated is not homogeneous within the workspace, $i.e.$, the field is stronger near the proximity of the powered magnets, the backward pulling force and forward pulling force vary depending on the location of the magnet. The needle tip pointing towards the electromagnets are closer than the electromagnets at the tail of the needle, then, the forward pulling force is emphasized. In the opposite situation, where the tail of the needle is closer to the electromagnets, the backward pulling force dominates.

In order to characterize this impact behaviour of the needle as a function of $D$ and $T$, we created an experimental setup with load cell (Transducer Techniques, GSO-1K) located inside of our magnetic system (Fig. \ref{fig:characterization}b). A two-dimensional sweep along $D$ and $T$ was implemented. We swept the range from 0.2 to 0.8 with 0.1 increments for $D$, and 50 ms to 250 ms with 50 ms increments for $T$. For each experiment, the impact motion lasted for 5 s resulting in more than 20 cycles. The force readings are used to 1) compute the average impact force generated, 2) compute the highest force density per time.

\paragraph*{Design Optimization}\mbox{} \\
Maximizing the impact force while keeping the length of the needle as miniature as possible is the main goal of the design optimization. Increasing the length of the magnet increases $\mathbf{F}_{m}$ and its mass, $m$. However, the travelling distance from tail to the tip of the needle decreases. Therefore, the optimization problem can be defined as 

\begin{equation}
    \begin{aligned}
    & \underset{l_{magnet}}{\text{maximize}}
    & & \mathbf{F}_{impact}(l_{magnet}) \\
    & \text{subject to}
    & & l_{magnet} \leq l_{tube}
    \end{aligned}
\end{equation}

\noindent where $\mathbf{F}_{impact}(l_{magnet})$ is the objective function, $l_{magnet}$ is the length of the magnet, $l_{tube}$ is the length of the tube. 

Defining $x_t = l_{tube} - l_{magnet}$ as the travelling length for a magnet from the tail to the tip of the needle and considering Eqs. (\ref{eqn:magnetic_force}), (\ref{eqn:force_eqn}) and (\ref{eqn:force_impact}), the impact momentum under a constant acceleration towards the tip can be defined as $l_{magnet} \sqrt{2ax_t}$. To maximize the impact force, this impact momentum should be maximized. Defining constants under a single constant parameter, $C = \rho_m \pi r_m^2 \sqrt{2a}$, yields the objective momentum function as $Cl_{magnet}\sqrt{(x_t)}$. Replacing $x_t = l_{tube} - l_{magnet}$ yields the optimization function as 

\begin{equation}
    \begin{aligned}
    & \underset{l_{magnet}}{\text{maximize}}
    & & f(l_{magnet}) = Cl_{magnet}\sqrt{(l_{tube} - l_{magnet})}.
    \end{aligned}
\end{equation}

\noindent $\frac{d\, f(l_{magnet})}{d\, l_{magnet}} = 0$ yields the optimum final relation as 

\begin{equation}
    l_{magnet} = \frac{2l_{tube}}{3}.
\end{equation}

\noindent By setting $l_{tube} = $ 19.05 mm for the needle proposed in this study, the optimum $l_{magnet}$ found to be 12.7 mm. The mathematical model that shows the relationship between the magnet length and tube length ratio vs. the impact force is also presented in Fig. \ref{fig:design}. A non-optimized design could result in a significant reduction of the maximum impact force that can be accomplished by the needle, which would make penetration infeasible as it can be seen in Fig. \ref{fig:design}.

\paragraph*{MagnetoSuture Setup}\mbox{} 

\begin{figure}
	\includegraphics[width=1\textwidth]{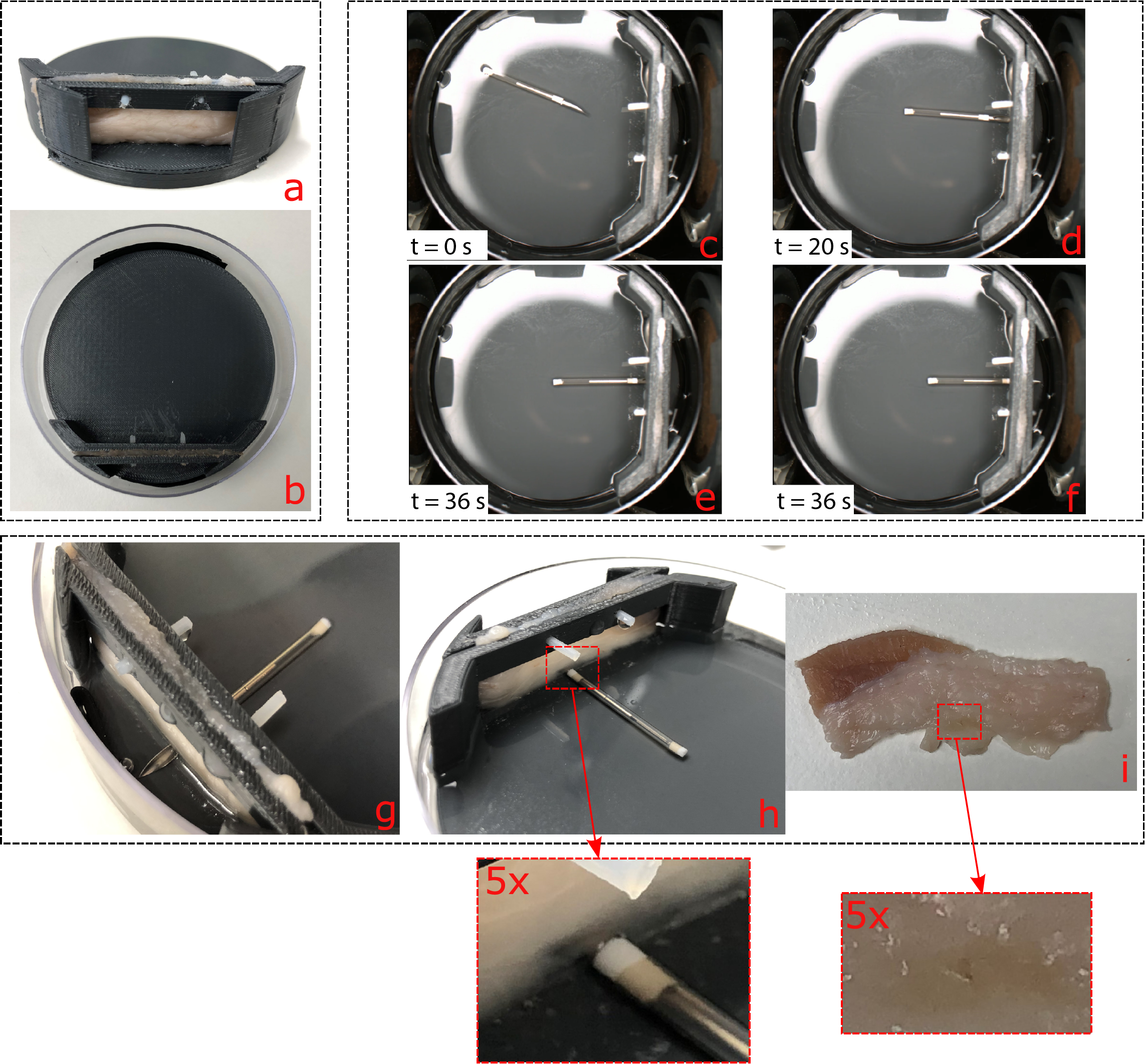}
	\caption{Suturing on a bacon strip: a-b) The experimental stage for bacon strip allows the needle and the piston magnet to remain at the center, where the magnetic pulling forces are stronger compared to the agar/gauze mesh experiments. The force characterizations are performed with the needle being at the center, which shows that the forces we acquire are on the required level for successful penetration. c-h) The needle is capable of penetrating more than 5 mm into the bacon strip. i) The minimally invasive needle penetration leaves a small difficult to see mark on the bacon.}
	\label{fig:bacon}
\end{figure}

In this work, we implemented an optimized magnetic hammer needle with the configuration shown in Fig.~\ref{fig:design}. To demonstrate the performance of the magnetic hammer needle, we employed our physical Magnetosuture system that was previously presented in~\cite{mair2020magnetosuture} and illustrated in Fig.~\ref{fig:concept}C. The needle was submerged in a viscous medium made by water-glycerol mixtures inside a Petri dish (diameter = \SI{85}{mm}). Tissue holders, as shown in Fig.~\ref{fig:gel} and Fig.~\ref{fig:bacon}, were customized for various tissue penetration tests. The magnetic hammer needle was tele-manipulated through external magnetic fields generated by an array of four uniformly spaced cylindrical electromagnets (EM). Each electromagnetic coil was made by approximately 12 wound layers of 54 turns of AWG 16 polyimide-coated copper wire (\(N_{em} = 12\times54\)). The inner diameter of the EM is \SI{85}{mm}, their outer diameter is \SI{98}{mm} (average diameter \(2\rho_{em} = 91.5\) mm), and their length is \(\ell_{em} = 60\) mm, as shown in Fig.~\ref{fig:concept}C. Four identical iron cores with diameters of \SI{52.18}{mm} and lengths of \SI{66.3}{mm} are inserted in the electromagnetic coils for boosting the magnetic field. The electromagnets are driven by two dual channel H bridge motor controllers (RoboClaw, Basic Micro Inc.) powered by an AC/DC converter capable of supplying \SI{62.5}{A} and \SI{48}{VDC} (PSE-3000-48-B, CUI Inc.). Visual feedback of the needle pose in the Petri dish was obtained by using a FLIR Blackfly camera (BFS-U3-13Y3C-C) with a resolution of 1280 $\times$ 1024 pixels. The workspace was illuminated by a ring light mounted on a custom 3D-printed adapter. 

\paragraph*{Teleoperation System with a Joystick}\mbox{} \\
The joystick inputs determine the direction of the needle. The joystick input is also being used to apply the sequence of pulling and pushing the magnetic piston along this direction based on the parameters $D$, $T$, $K_b$, and $K_f$. $K_b$ and $K_f$ values can also be tuned during the operation of suturing. Typically, $K_f$ is 1 to provide the maximum forward impact force. However, depending on the location of the needle, backward and forward pulling forces vary due to the distance to the powering electromagnets. Even though the same electrical current values are applied, the forces along backward or forward directions change during the operation. This situation becomes critical, especially if the suturing direction occurs at the end furthest away from the forward direction pulling coil. In this case, since the needle is closer to the backward pulling coil, the backward force dominates and the needle can not go forward, but it goes backward instead. To prevent such situations, tuning of $K_b$ is required, and this tuning was accomplished using the joystick during operation. Typically, $K_b$ is kept in the range that provides the oscillatory hammering behaviour while the net motion of the needle remains forward. 

\paragraph*{Force Characterization Measurements}\mbox{} \\
A single axis load cell (Transducer Techniques, GSO-1K) is located in the workspace of the magnetosuture setup. The voltage readings from the load cell's strain gauge is transferred to an Arduino microcontroller. Custom Python code is used to record the voltage readings in the form of the Arduino's digitized values. Load cell characterization experiments have shown that the measurements have the sensitivity of 5.69 mN. The tubular structure of the needle body is attached perpendicularly to the sensing region of the load cell by using a cyanoacrylate-based adhesive.

\paragraph*{Agar Gel with Gauze and Bacon Strip Penetration Force Characterization}\mbox{} \\
To characterize the penetration forces in the agar gel phantom tissue with gauze and bacon strips used in our experimental study, a needle penetration force recording system was setup by using a syringe pump (PHD ULTRA\textsuperscript{TM}, Harvard Apparatus) as a linear motion stage. A 12 G needle was attached on a single axis load cell (Transducer Techniques, GSO-1K), which was fixed on the moving part of the syringe pump. A 3D-printed tissue holder was placed along the needle's moving direction. For enabling repeated needle penetration tests on the same piece of tissue sample, the location of the tissue holder can be adjusted on the plane that is perpendicular to the needle's moving direction. One agar gel sample and three bacon strips samples were used for this study. We repeated the penetration on the same sample 9 times at various locations in order to generate penetration force ranges of both sample types. As a result of these experiments, we found out that average penetration force to the Agar gel with gauze is 248 mN with a standard deviation of 98 mN. Similarly, penetration into a bacon strip has the force requirements of 456.2 mN in average with 114.7 mN standard deviation. Each characterization experiment is repeated 5 times.

\paragraph*{Experimental Tissue Preparation}\mbox{}\\ 
A 1/16-inch (1.5875 mm) thick regular-cut bacon strip was used for the \textit{in-vitro} tissue penetration experiment. 0.5\% agarose gel with 3 mm thickness covered with a gauze mesh (CVS, Latex-free Gauze 5 CT) was used for the suturing experiment. Two similar testbeds were designed for the two experiments, as shown in Fig. 3 and Fig. 4, respectively. Both testbeds aimed to clamp samples perpendicularly to the planar needle workspace. Each testbed consisted of 1) two 3D printed pieces which could form a circular horizontal platform (radius = 41 mm) and a standing clamping frame (height = 11 mm), 2) a petri dish (inner radius = 43 mm), and 3) vegetable glycerin and water mixture (30 ml) (Glycerin Vegetable, Sanco Industries, Inc., Fort Wayne, IN). The major difference of the two testbeds was the 3D printed clamping frame position with respect to the circular platform. The clamping frame was posited near the petri dish walls (distance d=10 mm) and at the center of the petri dish for the first and second experiments, respectively. Hence the experimentally measured force was matched to the tissue penetration study, while adequate space was left for steering the needle in the suture study. For the first experiment for which a thinner sample was used, non-sticky tapes were inserted between 3D printed platform edges and petri dish walls to close the gap and enhance the clamping functionality. Additionally, two nylon sticks were used as press-fitting pins to provide extra clamping force on the top edge of the bacon strips.

\clearpage
\bibliography{reference}
\bibliographystyle{Science}

\paragraph*{Acknowledgments:}
Authors thank Wei-Hung Jung for his assistance on agarose gel preparation, Matt Shaeffer for his guidance in high speed camera usage, and Justin Opfermann for his various assistance and discussions. 

\paragraph*{Funding:} 
Research reported in this paper was supported by National Institute of Biomedical Imaging and Bioengineering of the National Institutes of Health under award numbers R01EB020610. The content is solely the responsibility of the authors and does not necessarily represent the official views of the National Institutes of Health.

\paragraph*{Author Contributions:} \mbox{} \\
Conceptualization: OE, XL, LM, YD, AK\\
Methodology: OE, XL, JG, LM, YB, AK\\
Software: OE, YB\\
Visualization: OE, JG\\
Data Curation: OE\\
Optimization Study: OE\\
Needle Manufacturing: OE, LM\\
Funding acquisition: AK\\
Supervision: YD, AK\\
Writing – original draft: OE, XL, JG, LM, YB, AK\\
Writing – review and editing: OE, XL, LM, YB, AK\\

\paragraph*{Competing interests:} Authors declare that they have no competing interests.
\paragraph*{Data and materials availability:} All data are available in the main text or the supplementary
materials.

\section*{Supplementary materials}
Materials and Methods\\
Supplementary Video 1: Impact-based needle for Magnetic Suturing

\end{document}